\documentclass[11pt]{article}

\usepackage[margin=1in, paperwidth=8.5in, paperheight=11in]{geometry}
\usepackage[inline]{enumitem}
\usepackage[hang,flushmargin]{footmisc}
\usepackage{multirow}
\usepackage{blindtext}
\usepackage{graphicx}
\usepackage{caption}
\usepackage{subcaption}
\usepackage{hyperref}
\usepackage{xcolor}
\newcommand{\link}[1]{{\color{blue}\href{#1}{#1}}}

\begin{document}

\setlength{\parindent}{0pt}
\setlength{\parskip}{4pt}
\rule{\linewidth}{1pt}
\begin{center}
\Large \textbf{Approximating Poker Probabilities with Deep Learning} \\
\end{center}
\rule{\linewidth}{1pt}

\vspace*{2mm}

\begin{center}
\textbf{Brandon Da Silva}\\
\vspace*{3mm}
\end{center}

\begin{abstract}

Many poker systems, whether created with heuristics or machine learning, rely on the probability of winning as a key input. However, calculating the precise probability using combinatorics is an intractable problem, so instead we approximate it. Monte Carlo simulation is an effective technique that can be used to approximate the probability that a player will win and/or tie a hand. However, without the use of a memory-intensive lookup table or a supercomputer, it becomes infeasible to run millions of times when training an agent with self-play. To combat the space-time tradeoff, we use deep learning to approximate the probabilities obtained from the Monte Carlo simulation with high accuracy. The learned model proves to be a lightweight alternative to Monte Carlo simulation, which ultimately allows us to use the probabilities as inputs during self-play efficiently. The source code and optimized neural network can be found at \link{https://github.com/brandinho/Poker-Probability-Approximation}.

\end{abstract}

\section{Introduction}

Poker has been an important area of research for Artificial Intelligence (AI) because of the complexity of the game environment. Partial observability means that agents must learn how to reason in the face of deception, while also learning how to deal with the non-deterministic nature of the environment. The specific variant of poker that has been used as the primary benchmark for imperfect information games is heads-up no-limit Texas Hold'em (HUNL). The standard version of HUNL used in the AI community has $10^{161}$ different decision points \cite{hunlstatesize}, making the state space comparable to Go, which has $10^{170}$ \cite{gostatesize}. 

Generally speaking, poker agents can broadly be bucketed into one of two categories:

\begin{enumerate}
\itemsep0em
\item Playing an approximate Nash Equilibrium strategy
\item Using opponent modelling to exploit behavioural biases
\end{enumerate} 

Irrespective of which bucket your agent falls into, we believe the following two probabilities (``the probabilities") are important inputs to use when designing poker agents\footnote{When approximating these probabilities we remain agnostic to our opponent's policy i.e. our opponent holds a random hand that is played out to the river. This reduces the size of the state space down to $5.56\times 10^{13}$ \cite{hunlstatesize}}: \begin{enumerate*}
\item Probability of winning the hand
\item Probability of a tie
\end{enumerate*}. We can approximate these probabilities with Monte Carlo simulation \cite{montecarlosimulation}. If the number of simulations, $N$, is large enough it will provide us with a tight approximation of the true probability. However, depending on the implementation, this can be quite computationally expensive to run during training. Under our current implementation it takes $0.46563$ seconds, on average, to run a simulation where $N=1000$.\footnote{We implemented a na{\"i}ve hand evaluator in python. Caching is not used because of improbable repeat rollouts} This makes training an agent over millions of hands infeasible. Alternatively, we can use a lookup table to evaluate hands quickly, which would significantly speedup the simulation. A popular implementation is the TwoPlusTwo evaluator, however it has 32.5 million entries with a total size of $\sim$123 MB\cite{twoplustwo}. We are presented with a clear tradeoff between time and memory. As a result, a neural network is used to approximate the approximated probabilities generated from the Monte Carlo simulation. Although approximating an approximation might raise some concerns, we are comfortable with this method due to the stability of the underlying Monte Carlo approximations. The neural network is able to approximate the probabilities with high accuracy and ultimately able to reduce the average inference time down to $0.00078$ seconds,\footnote{This includes the time it takes to calculate all the features in the input vector} which is a speed-up of $\sim$600x when compared to the na{\"i}ve implementation. Additionally, we now only need to store the weight matrices which take up 8.4 KB of memory, which is a reduction of $\sim$14,600x when compared to the large lookup table.

\section{Related Work}

Although the majority of poker papers are written about agents that learn an optimal policy with reinforcement learning (RL), there are a few poker papers that are specifically written about probability estimation. One paper outlined various methods for hand odds evaluators and ranking algorithms that can be used to estimate the probability of winning \cite{pokerestimatorportugal}. However, no evidence was provided to suggest these algorithms are good estimators. \\

Average Rank Strength (ARS) is a technique that was proposed to quickly compute poker abstractions, which are represented as the probability of having the best hand if the game reaches a showdown \cite{pokerspeedupportugal}. ARS makes use of three pre-computed 10 MB lookup tables to quickly compute these abstractions. While this method is fast and accurate, it still requires $\sim$3,500x more memory than our neural network implementation. \\

Another paper used a Support Vector Machine (SVM) classifier to determine whether or not you will win the hand. They then used the confidence score from the classifier as an estimate of the probability of winning \cite{pokersvm}. Even though they showed that their classifier achieved precision of $>60\%$, they did not show that the confidence score obtained from the SVM is a good approximation of the probability of winning a hand. Additionally, they take opponent actions as inputs into their model, while we don't. We remain agnostic to our opponent's policy because we believe it is important to model probabilities and opponents separately.

\section{Method}
	\subsection{Monte Carlo Simulation}
	Monte Carlo simulation is an effective solution when faced with intractable problems. It can be used to approximate the probability of having the winning hand by running a simulation over $N$ random scenarios, while holding your hole cards (two cards in your hand) and the current board cards constant. If $N$ is sufficiently large, then we will see the probability converge, which is an indication that the simulation provides a good approximation. We want a value for $N$ such that it approximates the probability well, yet at the same time it cannot be too large that it is infeasible to run millions of times.  \\
	
	We found that $N=1000$ is able to approximate the true probability within $\pm 2\%$,\footnote{Tightness of approximations are consistent across betting rounds despite the varying number of possible rollouts} which we believe is sufficient to use as an input for poker agents. We ran the simulation 25 times for each of the examples shown in Figure 1, and we see the approximations are both stable and tight. We use $N=100,000$ to simulate the true probability because we found that after running it ten times, all approximations fell within $0.4\%$ of each other. \\

	\begin{figure}[!htb]
	\centering
	\begin{subfigure}{0.5\textwidth}
		\centering
		\includegraphics[scale = 0.35]{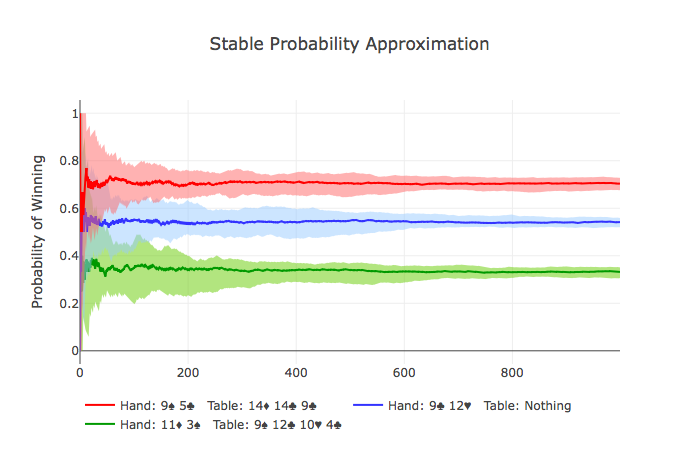}
		\captionof{figure}{Probability of Winning as $N \to 1000$}
	\end{subfigure}%
	\begin{subfigure}{0.5\textwidth}
		\centering
		\includegraphics[scale = 0.35]{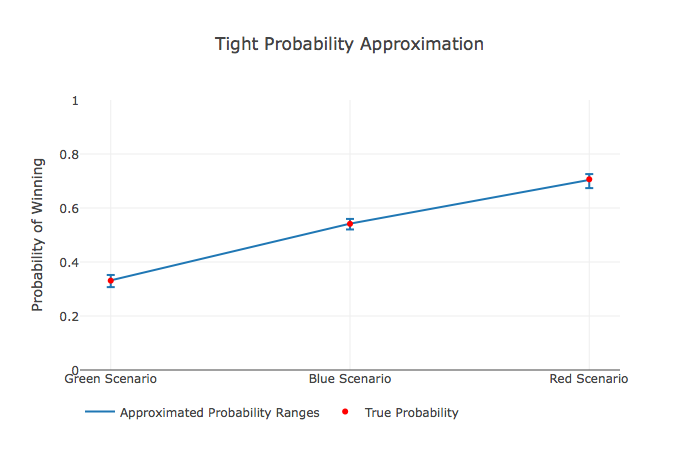}
		\captionof{figure}{Ranges obtained from Monte Carlo simulations}
	\end{subfigure}
	\caption{Monte Carlo simulation shows tight and stable probability estimates at N=1000}
	\end{figure}			
	
	Even though we believe $N=1000$ provides a good approximation, it takes 0.46563 seconds to run, which is too slow to use during training. Assuming our agent only plays 1 million hands, and it runs the simulation once per hand, this adds over 129 hours to training. Considering we need more than 1 million hands to train our agent, and the fact that we need to run the simulation once per betting round, rather than once per hand, it becomes evident that we need a better approach. We can trade off memory efficiency for time efficiency by using a large lookup table to evaluate hands, making Monte Carlo simulation a feasible approach. However, we will now be unable to implement our agent in memory-constrained environments, such as mobile applications. To get both time efficiency and memory efficiency, we use a neural network as a function approximation for the probabilities.
	
	\subsection{Dataset}
	To train the neural network we built a dataset of $250,000$ instances, in which the labels are the approximated probabilities from the Monte Carlo simulation. Inputs include a mixture of binary and continuous representations of hand/board states, as well as the calculated probabilities of achieving certain hands (Figures 2 \& 3). Binary variables were used to represent whether or not certain hands have been obtained with our hole cards or the board cards. Additionally, continuous variables were used to represent the rank of our hole cards, the number of suited cards, and the number of cards needed to complete a straight. A function was created using combinatorics to calculate the probabilities of getting each major hand given our hole cards and the cards on the board. Typically, Royal Flush is called out separately from Straight Flush, but we grouped them as one hand for our inputs. Lastly, we do not calculate the probability of achieving a High Card since all possible hands have a rank that is $\geq$ High Card. \\ 
	
	\begin{figure}[!htb]
	\centering
	\begin{tabular}{l|l}
	
	\textbf{Poker Hands} & \textbf{Description} \\ \hline
	Straight Flush & Five cards in a sequence, all the same suit \\ 
	Four of a Kind & All four cards of the same rank \\
	Full House & Three of a kind with a pair \\
	Flush & Any five cards of the same suit, but not in a sequence \\
	Straight & Five cards in a sequence, but not of the same suit \\
	Three of a Kind & Three cards of the same rank \\
	Two Pair & Two different pairs \\
	Pair & Two cards of the same rank \\
	High Card & When you have not made any of the hands above, the highest card plays
	
	\end{tabular}	 
	\caption{Possible Poker Hands}
	\end{figure}
		
	\begin{figure}[!htb]
	\centering
	\includegraphics[scale = 0.4]{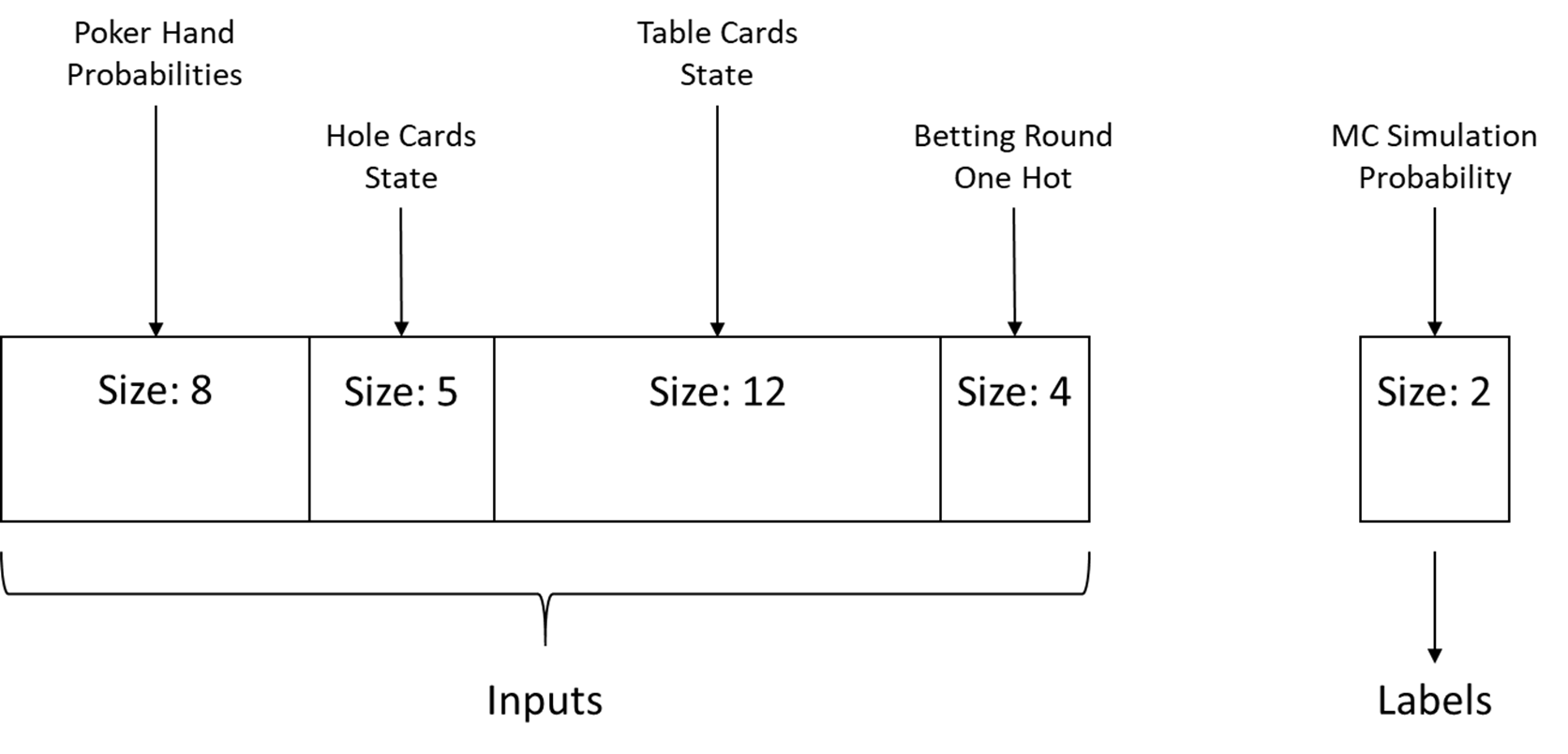}
	\caption{Dataset Structure}
	\end{figure}		

	\subsection{Deep Learning}
	Using the dataset described above, we train a fully-connected neural network to approximate the probabilities \cite{neuralnetworkbackprop}. The architecture for the network is $p$-24-12-$k$, where $p=29$ is the dimensionality of the input vector, and $k=2$ is the number of probabilities that we are predicting. The activation function used in the hidden layers is an Exponential Linear Unit (ELU) \cite{elu}, and the activation function used in the output layer is a sigmoid. The network uses Adam optimization to minimize Mean Squared Error (MSE) \cite{adam}. Training was done over $10,000$ epochs with batch sizes of $250$, and a train and test split of 90/10. Data was shuffled between each epoch to improve generalization. \\ 

	Although our model generalizes well, based on Figure 4 we believe that generalization can improve further as we increase the number of instances in our dataset. However, improved generalization is not the priority going forward. As we will see in the next section, expanding the input vector will likely have a higher impact on model performance. 

	\begin{figure}[!htb]
	\hspace*{1cm}\centerline{\includegraphics[scale = 0.45]{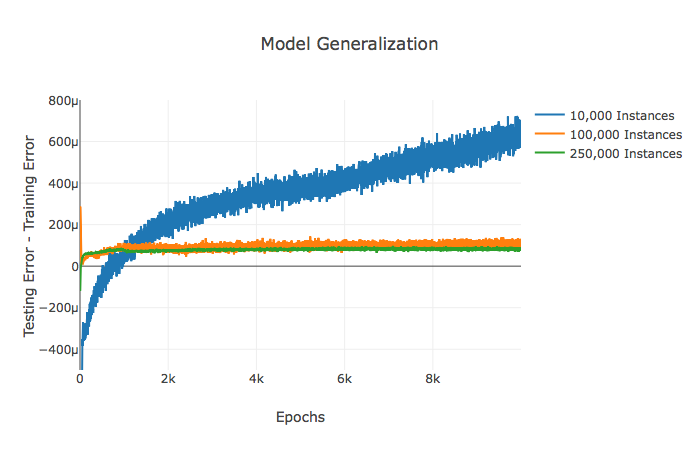}}
	\caption{Generalization improves as we increase the number of instances in our dataset}
	\end{figure}

\section{Experimental Results}
	The results on both the training and testing set can be seen in Figure 5. The top portion of the table shows the percentage of total predictions that fall within a certain deviation to the labels in which they are trying to predict. The bottom row of the table shows the Mean Absolute Error (MAE).  \\
	
	\begin{figure}[!htb]
	\begin{center}
	\begin{tabular}{c||c|c||c|c}
  	\multirow{2}{*}{Deviation} 
      	& \multicolumn{2}{|c||}{\textbf{Training Set} }
          	& \multicolumn{2}{c}{\textbf{Testing Set}} \\            
  	& Probability of Win & Probability of Tie & Probability of Win & Probability of Tie \\  \hline
  	Within 0.5\% & $14.75\%$ & $37.06\%$ & $14.60\%$ & $36.79\%$ \\      
  	Within 1.0\% & $28.25\%$ & $59.88\%$ & $27.98\%$ & $59.90\%$ \\  
  	Within 2.0\% & $49.82\%$  & $81.00\%$ & $49.44\%$ & $81.21\%$ \\      
  	Within 3.0\% & $63.93\%$ & $89.65\%$ & $63.58\%$ & $89.59\%$ \\  
  	Within 4.0\% & $72.95\%$ & $93.93\%$ & $72.29\%$ & $93.69\%$ \\      
  	Within 5.0\% & $79.14\%$ & $96.07\%$ & $78.57\%$ & $95.88\%$ \\  
  	Within 10.0\% & $93.62\%$ & $98.68\%$ & $93.06\%$ & $98.62\%$ \\      
  	Within 20.0\% & $99.34\%$ & $99.60\%$ & $99.18\%$ & $99.54\%$ \\ [5pt] \hline \hline
  	\textbf{MAE} & $3.33\%$ & $1.42\%$ & $3.44\%$ & $1.46\%$ \\ 
	\end{tabular}
	\end{center}
	\caption{The model is able to generalize well to the Testing Set}
	\end{figure}
	
	We see that the model is able to predict the probability of a tie more accurately than the probability of a win. However, the model is still able to predict the probability of a win within a 5\% deviation for over $\frac{3}{4}$ of the instances. The small deviation between the training and testing set show that the model generalizes well.  \\
	
	 We also experimented with a neural network that had a softmax output layer with $k=3$ neurons $[P(win) \quad P(tie) \quad P(lose)]$. We found the results to be almost identical to the current architecture we recommend, with the errors on the probability of losing mirroring the errors on the probability of winning. As a result, we decided to keep our current architecture where $k=2$. We assume an agent is able to impute the probability of losing given the probability of winning and the probability of a tie.  \\

We also looked at the worst prediction in both the training and testing set. Interestingly, both instances had strong hands on the board that usually would end in a tie. As a result, our model predicted a low probability of a win and a high probability of a tie. The example in the training set had a straight on the board (5, 6, 7, 8, 9), while we had (3, 10) in our hand. Thus, we had the higher straight. Similarly, the example in the testing set had a Jack-high flush on the board. We had a King of the same suit in our hand, giving us the higher flush. These scenarios are unable to be accurately represented given the current input vector. We do not have a binary variable that indicates whether or not the hand formed with our hole cards is better than the board alone. We believe expanding the input vector to incorporate this variable will solve this particular problem. 

\section{Conclusion}
The approach introduced in this paper is a promising start to efficiently approximating poker probabilities. We explored the use of deep learning for poker probability approximation. Neural networks were used as a means to reduce the memory requirement and execution time during inference, allowing RL agents to efficiently approximate the probabilities during self-play. Experimental results show that the model generalizes well, while also being able to approximate the majority of instances within $3\%$ of the labels. However, we believe that the results can further be improved by expanding our input vector and using a larger dataset.

\end{document}